%% file: paper.tex
\documentclass{article} % For LaTeX2e
\usepackage{nips15submit_e,times}
\usepackage{amsmath,amsfonts,amsthm,amssymb}
\usepackage{url}
\usepackage{xurl}
\usepackage{graphicx}
\usepackage{caption}
\usepackage{subcaption}
\usepackage{algpseudocode}
\usepackage{bbm}
\usepackage{float}
\usepackage{framed}
\usepackage{enumerate}
\usepackage{color}
\usepackage[utf8]{inputenc}
\usepackage{mathtools}
\usepackage[colorlinks=true, linkcolor=red, urlcolor=blue, citecolor=blue]{hyperref}
\usepackage[symbol]{footmisc}

\title{Understanding Why ViT Trains Badly on Small Datasets: An Intuitive Perspective}

\author{
Haoran Zhu\thanks{Equal contributors.}\\
New York University\\
\texttt{hz1922@nyu.edu} \\
\And
Boyuan Chen$^{*}$\\
New York University\\
\texttt{boyuan.chen@nyu.edu} \\
\And
Carter Yang\\
New York University\\
\texttt{py2097@nyu.edu}
}

\nipsfinalcopy % Uncomment for camera-ready version

\begin{document}

% \footnote[$^{\star}$]{\text{ Equal contributors.}}

\maketitle

\input{abstract}

% 1
\input{introduction}

% 2
\input{model_data}

% 3
\input{accuracy}

% 4
\input{visualization}

% 5
\input{representation.tex}

\input{conclusion}

\input{acknowledgement}

\bibliographystyle{unsrt}
\bibliography{reference}

\newpage

\newpage
\input{appendix}
\end{document}

%% file: abstract.tex
\begin{abstract}
Vision transformer (ViT) is an attention neural network architecture that is shown to be effective for computer vision tasks. However, compared to ResNet-18 with a similar number of parameters, ViT has a significantly lower evaluation accuracy when trained on small datasets. 
To facilitate studies in related fields, we provide a visual intuition to help understand why it is the case.
We first compare the performance of the two models and confirm that ViT has less accuracy than ResNet-18 when trained on small datasets. We then interpret the results by showing attention map visualization for ViT and feature map visualization for ResNet-18. The difference is further analyzed through a representation similarity perspective. We conclude that the representation of ViT trained on small datasets is hugely different from ViT trained on large datasets, which may be the reason why the performance drops a lot on small datasets. Our code and documentation are publically available at:  \url{https://github.com/BoyuanJackChen/MiniProject2_VisTrans}. 
\end{abstract}

%% file: introduction.tex
\section{Introduction}
Attention mechanism has become the most effective tool in natural language processing tasks. In recent years, it has been proven to perform well on computer vision tasks, such as image detection, image classification and video processing. With the advent of Visual Transformer (ViT)\cite{ViT}, the pure attention network began to rival the accuracy of convolutional neural networks (CNN) in image classification tasks. Further research on vision transformers will not only improve the capability of machine learning in vision tasks, but also improve our understanding on transformers and their relationship with CNN. 

Nonetheless, one major drawback of vision transformer is its bad performance on small-scale datasets \cite{naimi2021hybrid}. Traditional CNN can be trained to make high accuracy predictions on the test dataset, and their accuracy increases as we increase the number of parameters and layers. On the other hand, vision transformers usually have poor performance when trained on small datasets. Existing Methods such as Shifted Patch Tokenization (SPT) and Locality Self-Attention (LSA) are proven to improve the transformers' accuracy on small datasets\cite{ViT_for_small}, yet their accuracy is still lower than CNN's. 

We explore the reason why vision transformers perform worse than CNN on smaller datasets. We provide visual evidence, such as attention visualization and forward propagation, and representation similarity analysis. We expect our results may contribute to the understanding of the attention mechanism on image data, as well as to inspire a new solution to improve vision transformer networks.

The contributions of our work could be summarized as follows:
\begin{itemize}
\item By comparing the performance of ViT and ResNet on CIFAR-10, CIFAR-100, and SVHN datasets, we confirm that ViT does not perform well on small datasets compared with CNN.
\item We conduct attention visualization on ViT and feature map visualization on ResNet to visualize the weights of each layer in each model.
\item We empirically measure the representation similarity between ViT and ResNet on small datasets and compare the difference on large datasets in \cite{raghu2021do}. Unlike \cite{raghu2021do}, which mainly compares the representation similarity on large datasets, we focus on analyzing differences and explore reasons for performance drop of ViT on small datasets.
\end{itemize}

%% file: model_data.tex
\section{Model and Data}
 
We initialize a ViT model based on \cite{ViT} and train it on three image datasets: CIFAR-10, CIFAR-100, and Street View House Numbers (SVHN) \cite{netzer2011reading}. This is done with two primary objectives: to re-generate existing literature results on ViT for small datasets \cite{ViT_for_small, liu2021efficient}, and to understand which kind of small dataset ViT learns less well, by comparing the evaluation accuracy. 
 
We compare the performance of our ViT model, which has 9.6M parameters, to the performance of a standard ResNet18 model, which has 11.5M parameters. We choose the latter for comparison because it is widely used to assess model efficiency, and its parameter count is relatively close to that of ViT, when compared to other ResNet architectures. 
 
Overall, ViT performs comparably to ResNet on SVHN, but significantly worse on CIFAR-10 and CIFAR-100. We will discuss their respective accuracy and provide evidence to support this finding in the following sections.

\subsection{Dataset and Augmentation}
We train models for image classification using the CIFAR-10, CIFAR-100, and SVHN. The CIFAR-10 data contains 50k training images and 10k testing images with 10 classes, each class having the same number of images for both training and testing sets. The CIFAR-100 data has the same image size, and the same volumes of training and testing dataset. The only difference from CIFAR-10 is that CIFAR-100 has 100 classes, evenly assigned to images in both training and testing sets. Therefore, the number of samples in each class is only 1/10 of that in CIFAR-10, making the training harder. The SVHN dataset contains 600k images of digits of house numbers, and each label is the digit that image shows. All three datasets have images of 32×32 pixels in three channels of color. The unification of this factor eliminates the possible difference in outcome based on image size. 
 
We introduce data augmentation methods for image classification tasks including flipping and cropping. For the training set, we cropped the input image at a random location in 32 × 32 pixels with a padding of 4, and randomly flip the image horizontally with the probability of 0.5, which is applied to all types of datasets.
 
For both ResNet and ViT, we did not implement normalization on pixel values.

\subsection{Model Architecture}
For the ResNet, we implement the standard ResNet-18 architecture, as it is widely used for comparison in many works on image classification. Each residual block has 2 convolutional layers, with three expansions at a rate of 4 every two residual blocks. 
 
\begin{figure}[H]
\centering
\includegraphics[width=14cm]{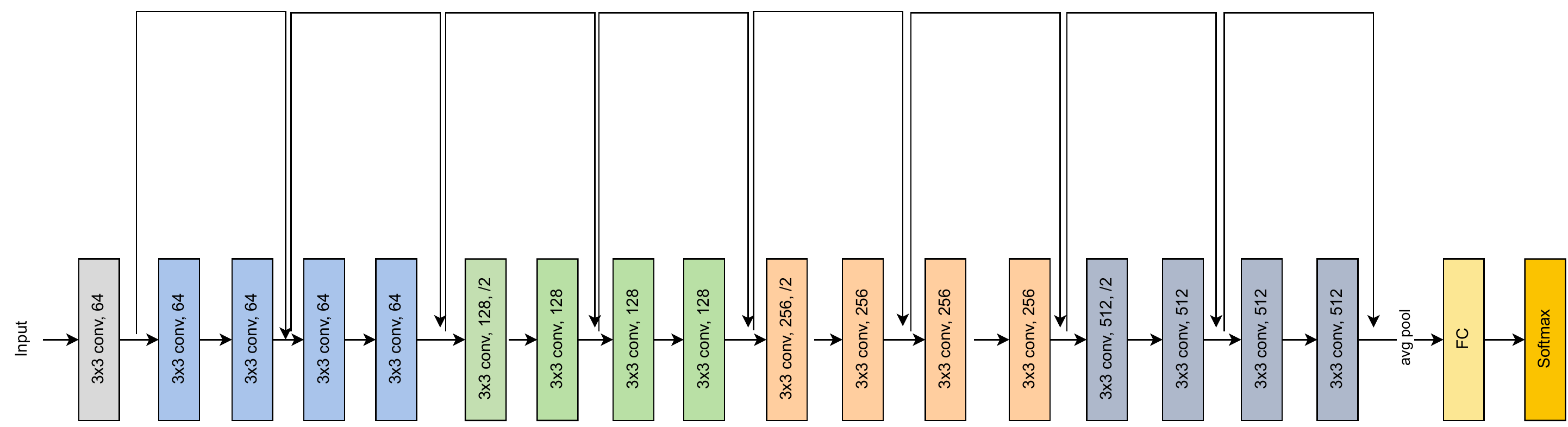}
\caption{ResNet-18 architecture.}
\label{fig:resnet18}
\end{figure}
 
For Vision Transformer (ViT), we divide the image into 4 batches. Each attention layer has 8 heads, each having a dimension of 64. 
The transformer encoder has a depth of 6, and a drop-out rate of 0.1.
Finally, the MLP layer has a dimension of 512, and a drop-out rate of 0.1. 
 
\begin{figure}[h]
\centering
\includegraphics[width=13cm]{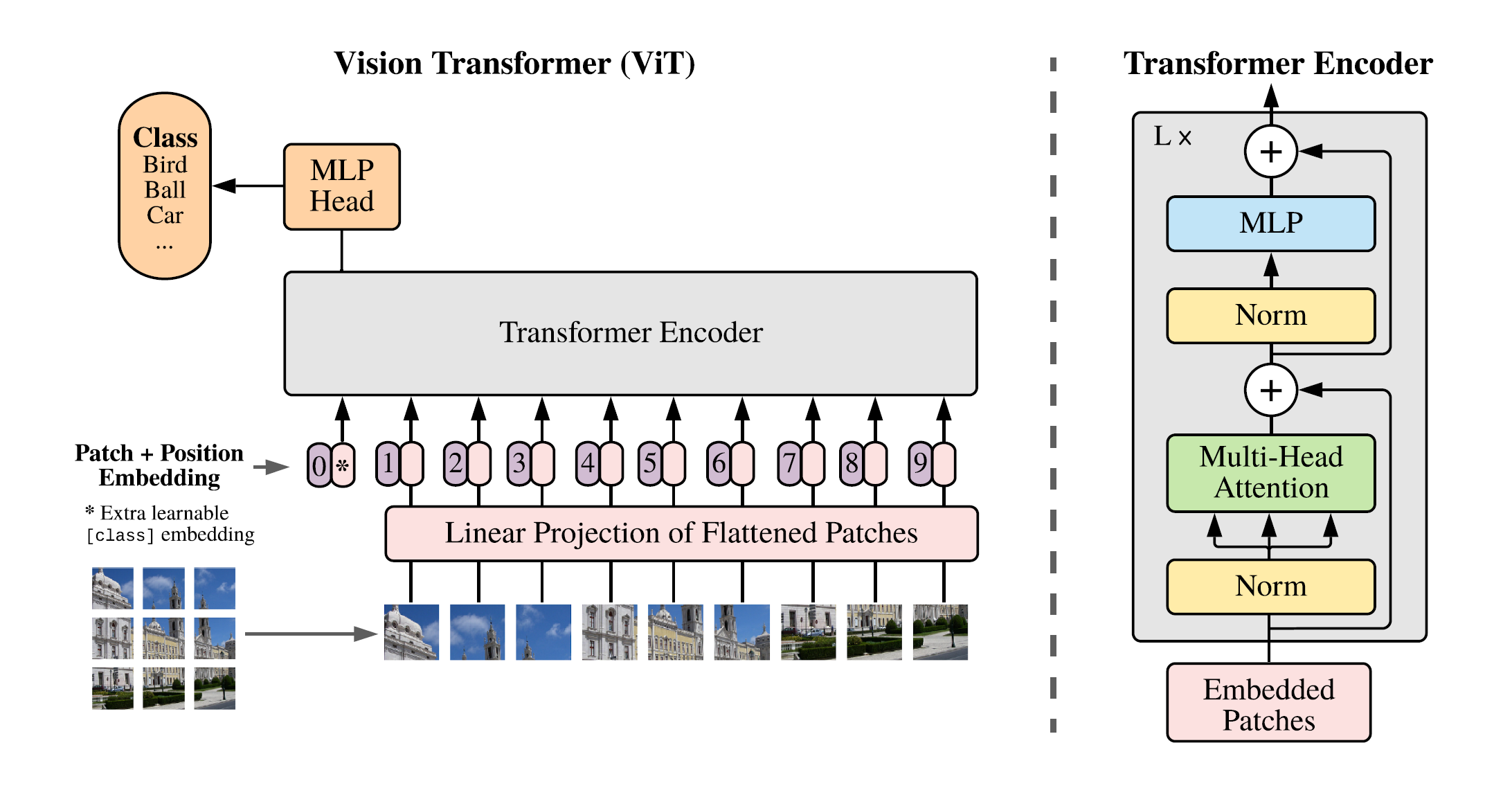}
\caption{ViT architecture, image from \cite{ViT}. In our experiments, the transformer encoder has L=6 layers. }
\label{fig:ViT structure}
\end{figure}

\subsection{Training Details}
We train two models: ResNet-18, and Vision Transformer, on 3 different datasets: cifar10, cifar100, and svhn. To make it fair, all of the hyper-parameters are kept the same such as learning rate = 1e-4, batch size = 100, and using adam optimizer. We ran each experiment for 500 epochs. and use wandb (Weights and Biases) \cite{wandb} library to track and visualize the results. The built-in visualization features in wandb provide multiple plots of metrics mainly about train/test loss and accuracy, allowing us to compare across different models with the same dataset.

%% file: accuracy.tex
\section{Model Accuracy}
\begin{figure}[!ht]
 \centering
 \begin{subfigure}[!ht]{0.6\textwidth}
     \centering
     \includegraphics[width=\textwidth]{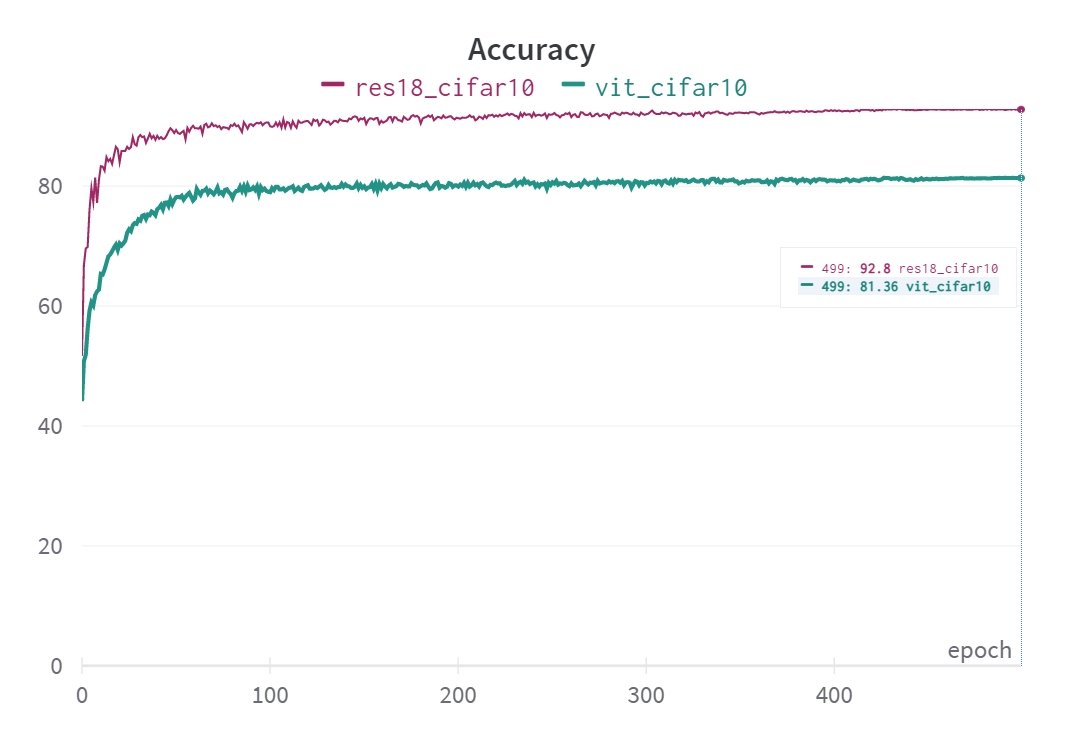}
     \caption{CIFAR-10 accuracy}
     \label{fig:cifar-10 results}
 \end{subfigure}

 \begin{subfigure}[!ht]{0.6\textwidth}
     \centering
     \includegraphics[width=\textwidth]{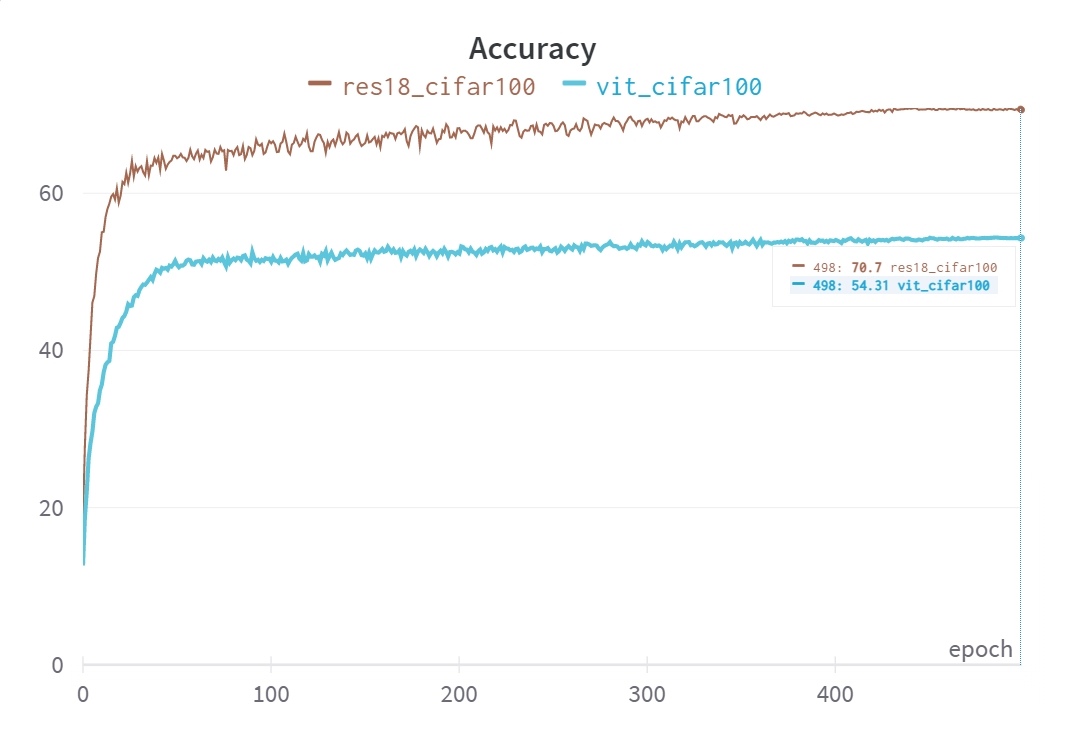}
     \caption{CIFAR-100 accuracy}
     \label{fig:cifar-100 results}
 \end{subfigure}

 \begin{subfigure}[!ht]{0.6\textwidth}
     \centering
     \includegraphics[width=\textwidth]{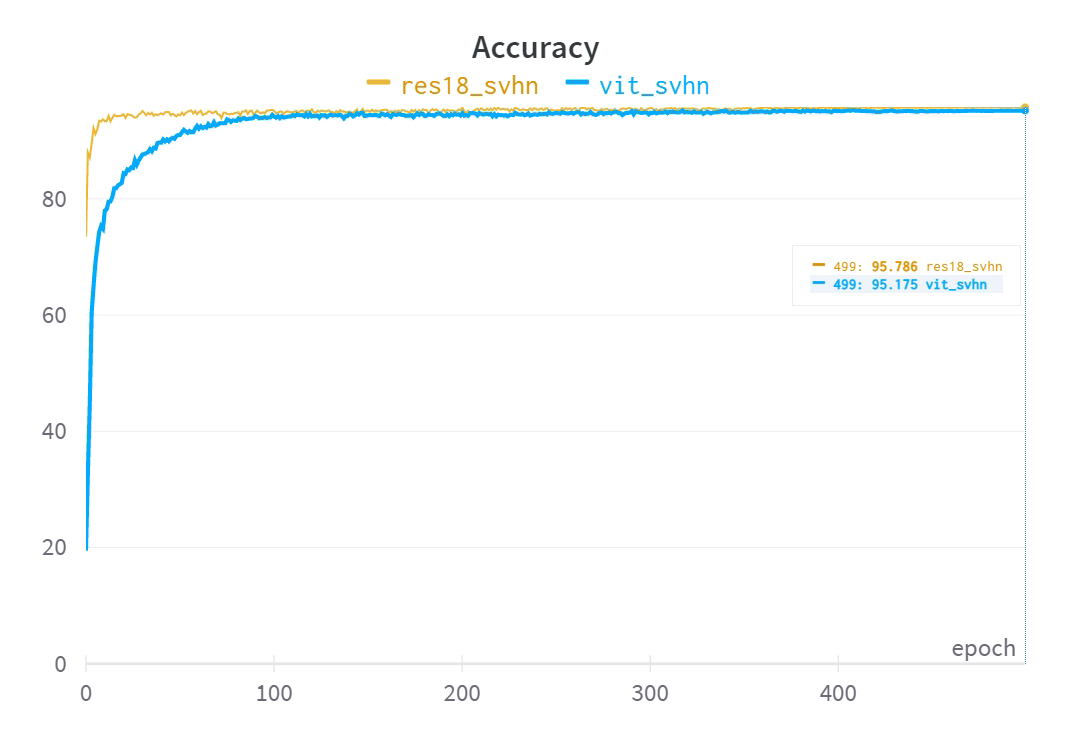}
     \caption{SVHN accuracy}
     \label{fig:svhn results}
 \end{subfigure}
    \caption{Accuracy in 500 epoch}
    \label{fig:Accuracy curves}
\end{figure}

Table \ref{tab:accuracy_table} shows the performance of ViT compared with ResNet18 on CIFAR-10, CIFAR-100 and SVHN dataset after training for 500 epochs. Figure \ref{fig:cifar-10 results}-\ref{fig:svhn results} show the accuracy testing curve during training. We can see that ViT performs significantly worse on CIFAR-10 and CIFAR-100 compared to ResNet18. The error rate of the former is twice of the latter. Nonetheless, ViT performs equally well on SVHN, a colored dataset on digit recognition, though its convergence is slower than ResNet from Figure \ref{fig:svhn results}.

This result confirms the assumption that ViT performs worse on small datasets. ViT achieves a similar result on SVHN because of the simplicity of the dataset. In general, ViT also performs well on MNIST, which is a one-channel version of digit recognition. If the model can fit well on one channel, then it is likely to also fit well on three channels. 

% Accuracy table
\begin{table}[!ht]
    \centering
    \begin{tabular}{|c|c|c|c|}
         \hline
         &  CIFAR-10 & CIFAR-100 & SVHN\\
         \hline
         ViT & 81.36 & 54.31 & 95.17\\
         \hline
         ResNet18 & \textbf{92.8} & \textbf{70.7} & \textbf{95.78}\\
         \hline
    \end{tabular}
    \caption{Top-1 accuracy(\%) of ViT and ResNet18, trained from scratch on different small datasets (500 epochs).}
    \label{tab:accuracy_table}
\end{table}

% Evaluation accuracy during training

%% file: visualization.tex
\section{Visualization of Layers}

In this section, we gain intuitions about what each layer of a model does by using attention weights extraction for ViT model and feature mapping for ResNet. Unfortunately, the two visualizations are not directly comparable, as they use vastly totally different learning strategies. However, the visualization tools can still help us understand the logic behind the black box of parameters. 

\subsection{Attention Weights for ViT}
Attention weights visualization is proposed in the original ViT paper \cite{ViT} to demonstrate how the model processes image data. As the method instructs, we first extract the attention layer with shape ($n_{heads}$, $n_{patch}+1$, $n_{patch}+1$) in each transformer block. Then we average the weight across the heads and then add with an identity matrix to account for residual connection. We then normalize and reshape the matrix to form a weight mask. To facilitate visualization, each weight mask is scaled by a common factor so that the max weight of all masks is equal to 1. Finally, each weight mask overlaps on top of the original image. The bright areas receive more attention, and the darker areas receive less attention. 
 
\begin{figure}[ht]
\centering

\begin{subfigure}[ht]{\textwidth}
    \centering
    \includegraphics[width=12cm]{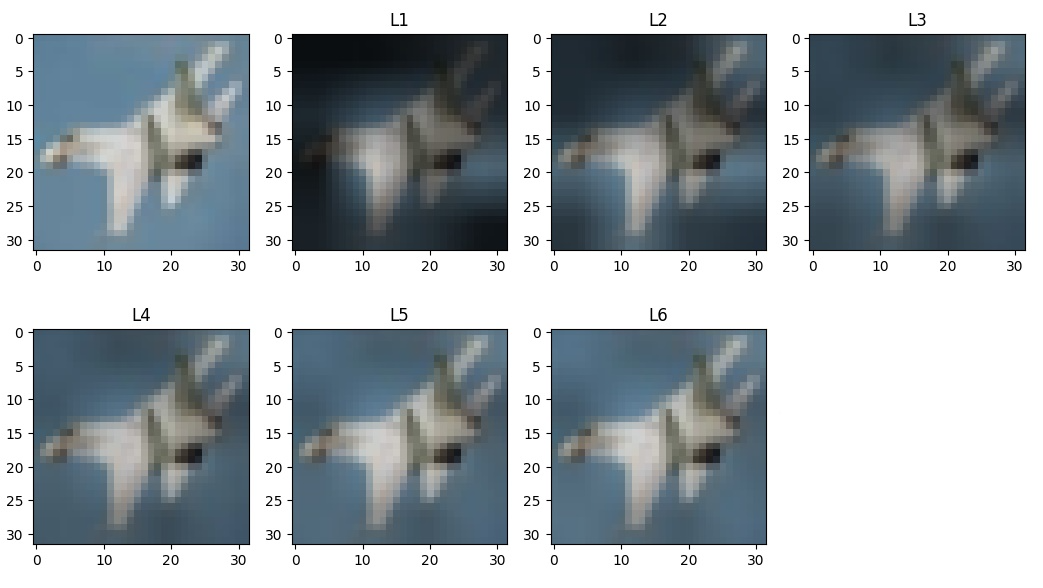}
    \caption{Image of airplane from CIFAR-10 dataset.}
    \label{fig:ViT_plane_vis}
\end{subfigure}

\begin{subfigure}[ht]{\textwidth}
    \centering
    \includegraphics[width=12cm]{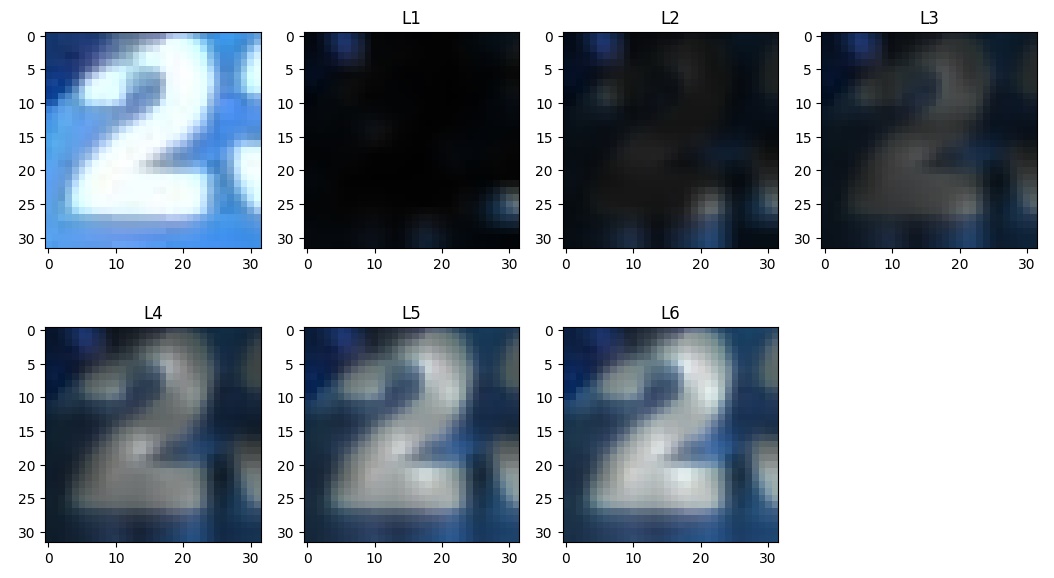}
    \caption{Image of number 2 from SVHN dataset.}
    \label{fig:ViT_2_vis}
\end{subfigure}

\caption{Attention weight visualization from ViT. The top two rows show the model trained on CIFAR-10, forwarding an image of an airplane; the lower two show the same model trained on SHVN, forwarding an image of number 2. The input is the 10th image from CIFAR-10, with the label of an airplane. The original image is shown on the top-left. The rest shows the original image overlapped with the weight mask of each attention layer. Bright regions represent higher attention weights (close to 1); darker regions represent lower attention weights (close to 0).}
\label{fig:ViT_vis_plane}
\end{figure}

While the original work only shows the weight mask for the first attention layer, here we provide the visualization of all the 6 layers' attention weights in Figure \ref{fig:ViT_vis_plane}. The first layer exhibits concentrated attention on a small area, while later layers expand their attention to the whole image. Our experiments show that the first layer tends to put more weight on areas with higher contrast in the neighborhood, and future layers expand their attention from the previous layer. 
 
Based on the visualizations, we speculate that the ViT model is trained to put more attention on regions with higher local contrast. While this strategy works well on most pictures in CIFAR-10 dataset, there are also images where this strategy does not perform well. Such examples can be found in the Appendix.

\subsection{Feature Map Visualization for ResNet}
A feature map is the output of a convolution layer in a CNN network. The values across channels are averaged, so we can directly visualize the output in grayscale. For a ResNet18 network, we can generate 17 feature maps. For dimensions, layers 1-5 are of size $(32, 32)$; layers 6-9 are of size $(16, 16)$; layers 10-13 are of size $(8,8)$; layers 14-17 are of size $(4,4)$. 
 
\begin{figure}[!ht]
\centering
\begin{subfigure}[ht]{\textwidth}
    \centering
    \includegraphics[width=14cm]{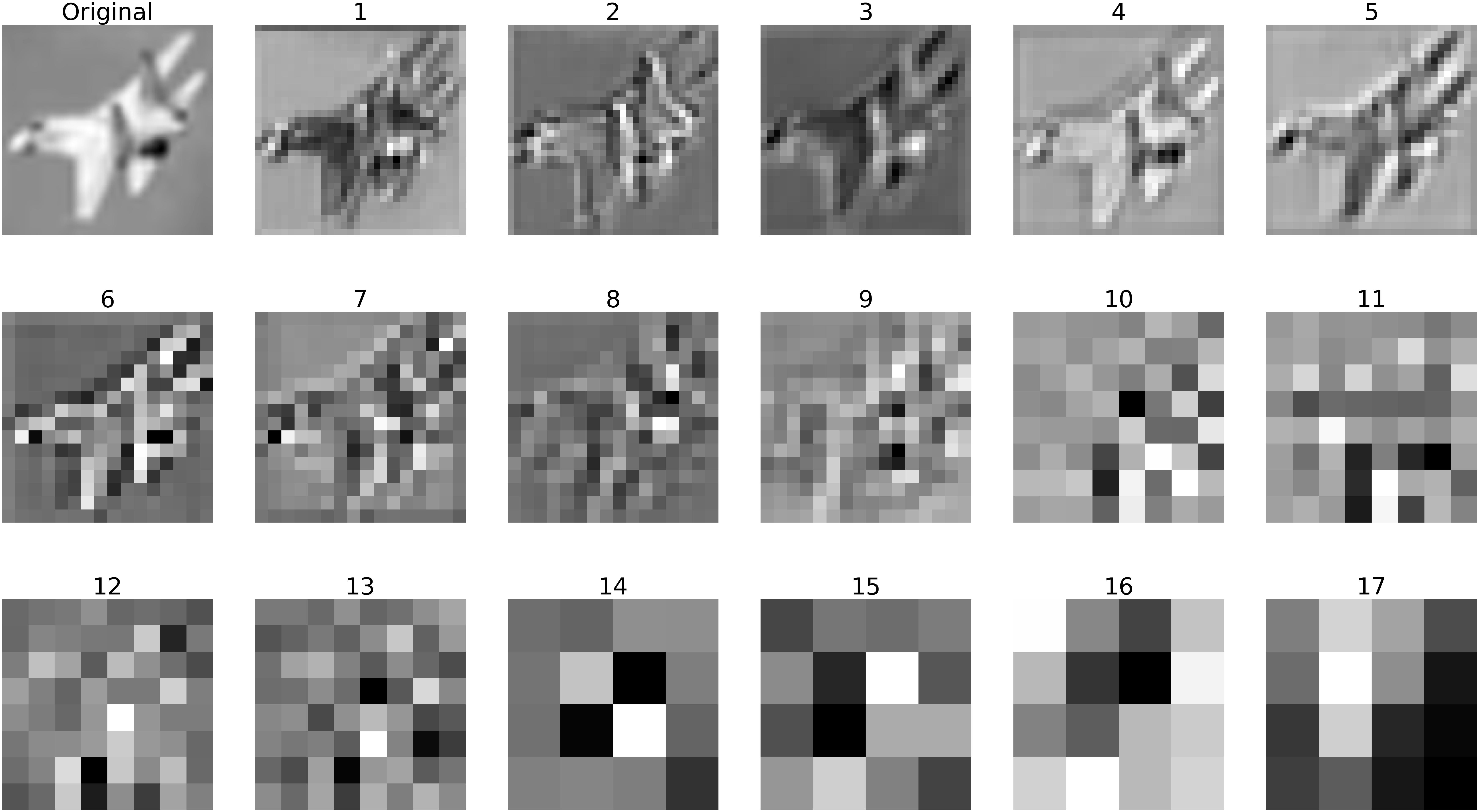}
    \caption{Image of number 2 from SVHN dataset.}
    \label{fig:ResNet_2_vis}
\end{subfigure}

\vspace{5mm}
\begin{subfigure}[ht]{\textwidth}
    \centering
    \includegraphics[width=14cm]{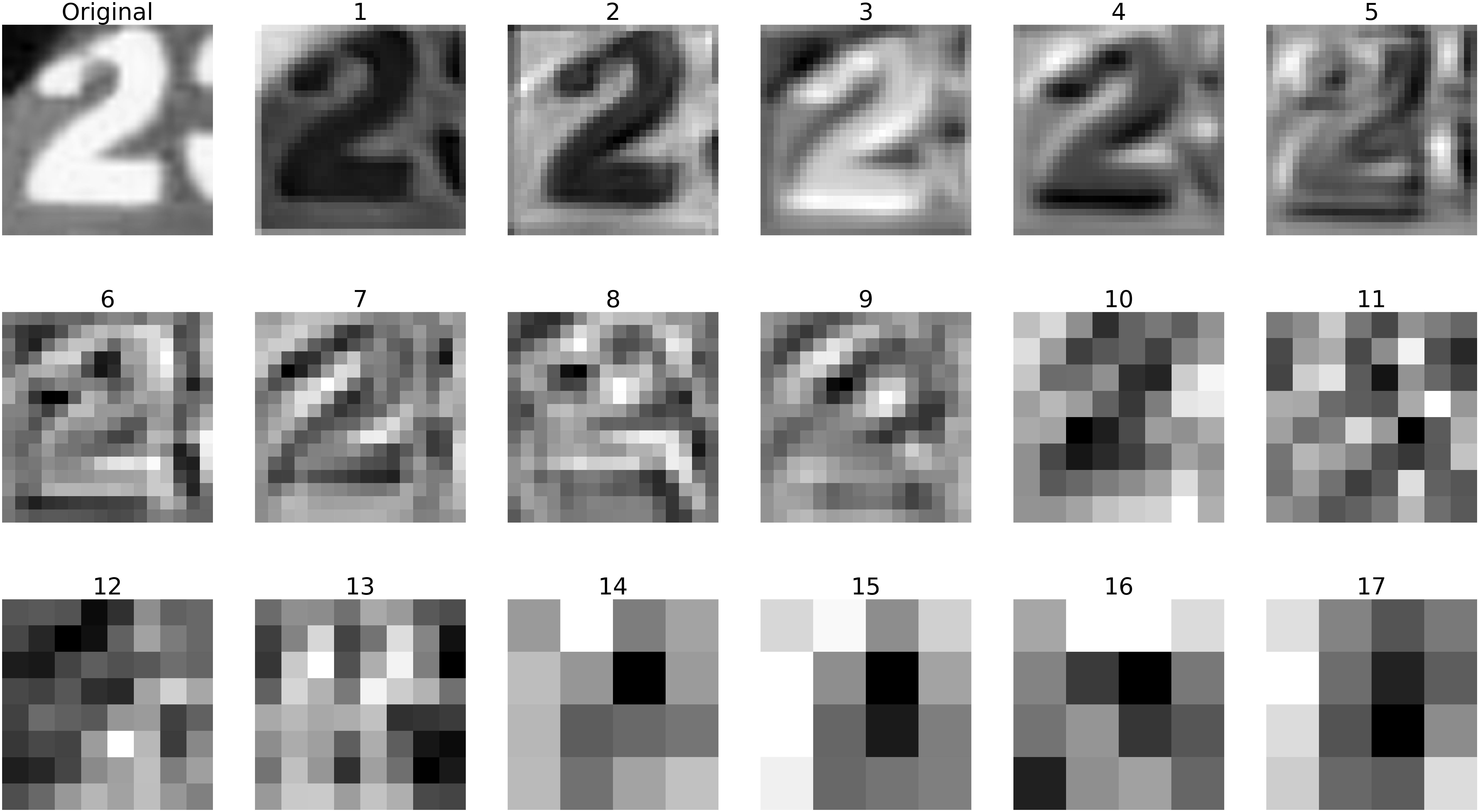}
    \caption{Image of number 2 from SVHN dataset.}
    \label{fig:ResNet_2_vis}
\end{subfigure}

\caption{Feature map visualization from lower layers to higher layers in ResNet18. The top three rows show the model train on CIFAR-10 forwarding an airplane image; the lower three rows show the same model train on SHVN forwarding an image of number 2. The top-left is the original image in grayscale; from left to right top to bottom, we exhibit feature maps of convolution layer 1 to convolution layer 17. }
\label{fig:resnet_vis_plane}
\end{figure}

Figure \ref{fig:resnet_vis_plane} shows the feature maps of the model forwarding the same images as we used for ViT. 
 
%% file: representation.tex
\section{Representation Similarity Analysis}
\begin{figure}[h]
\centering
% \begin{minipage}[b]{0.7\textwidth}
\includegraphics[width=10cm]{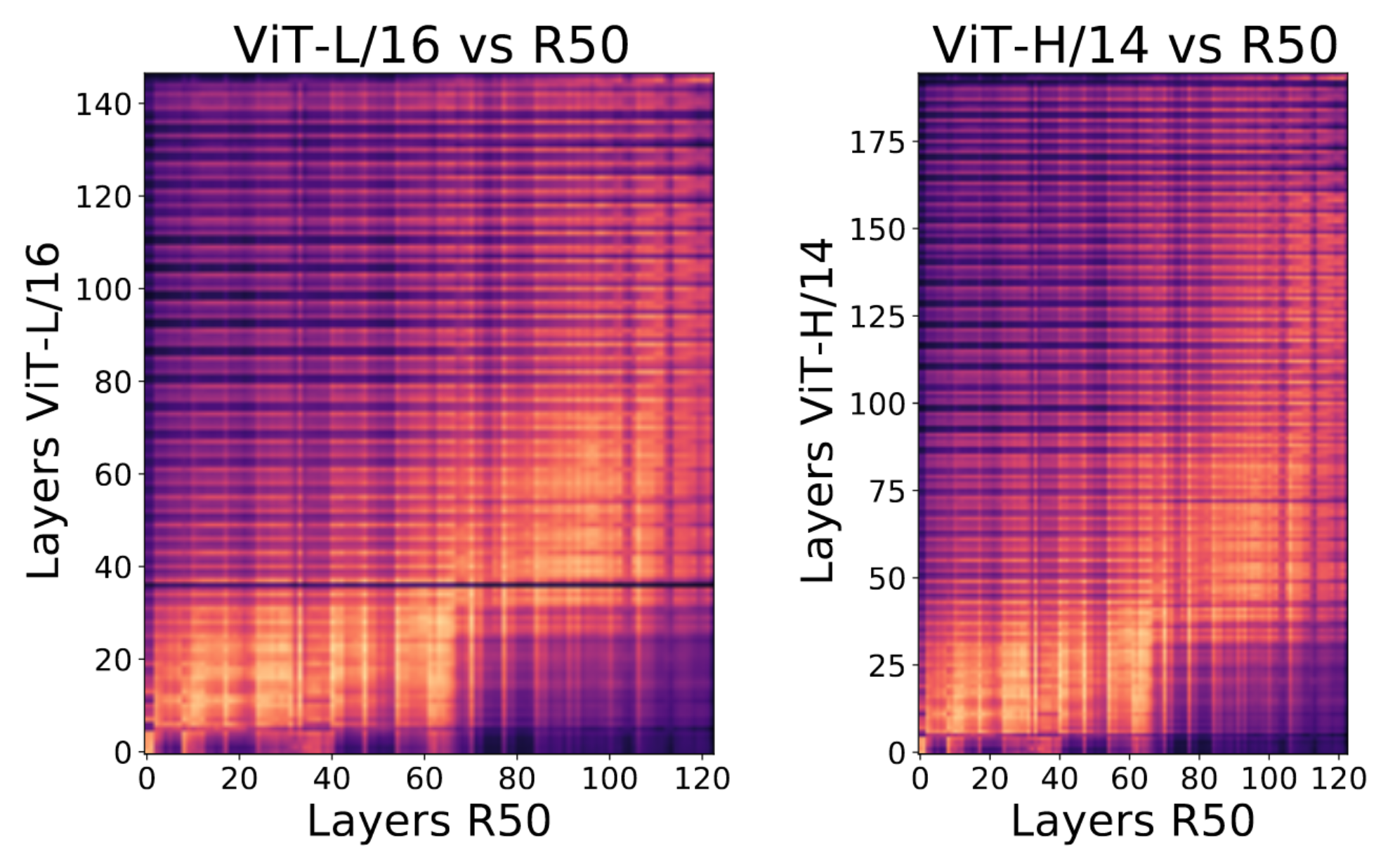}
\caption{Figure from \cite{raghu2021do} representation similarity between ViT and ResNet on large datasets (JFT-300M)}
\label{fig:cka_jft_300m}
% \end{minipage}
\end{figure}

\begin{figure}[h]
 \centering
 \begin{subfigure}[t]{0.3\textwidth}
     \centering
     \includegraphics[width=\textwidth]{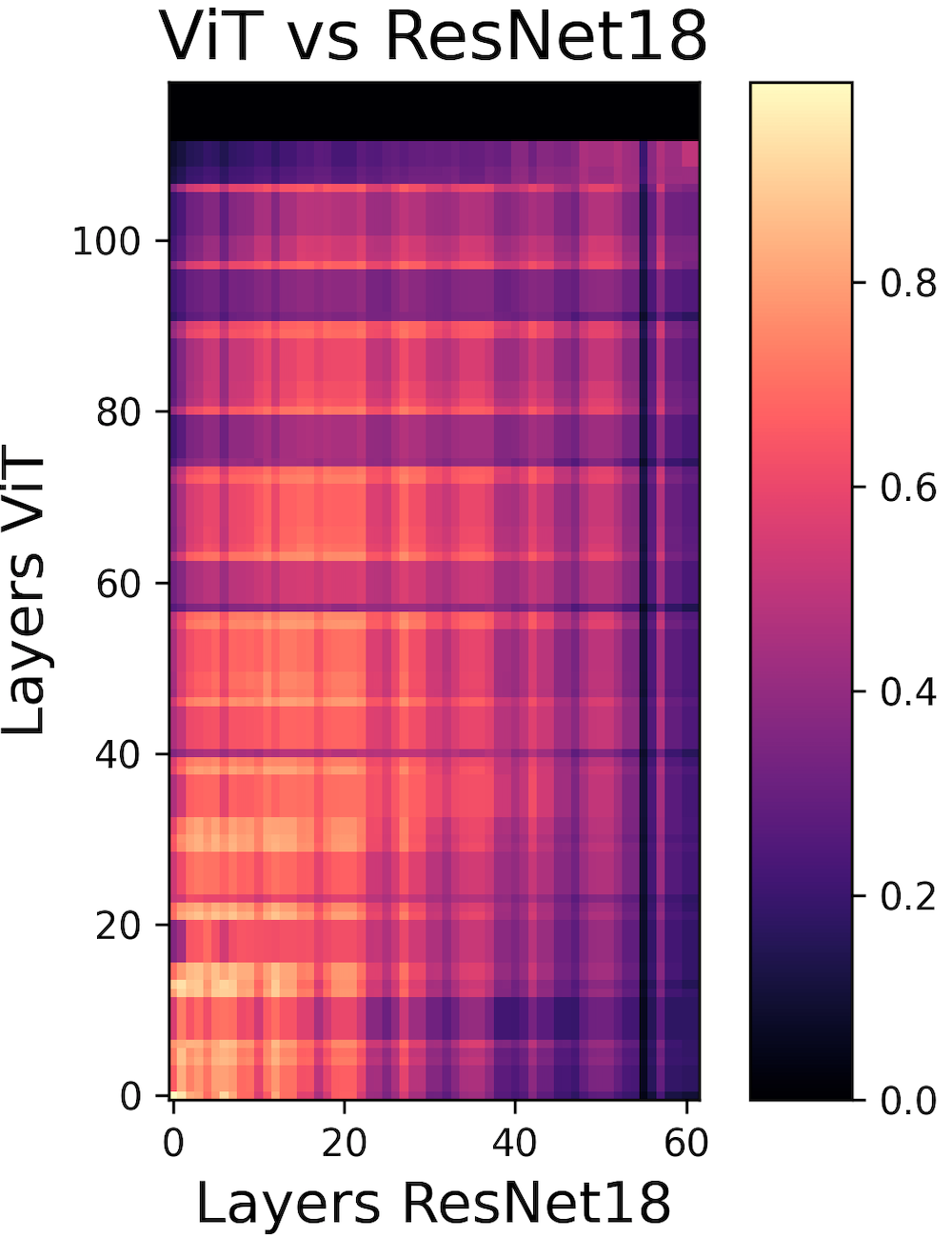}
     \caption{Representation similarity between ViT and ResNet on CIFAR-100}
     \label{fig:cka_cifar100}
 \end{subfigure}
 \hfill
 \begin{subfigure}[t]{0.3\textwidth}
     \centering
     \includegraphics[width=\textwidth]{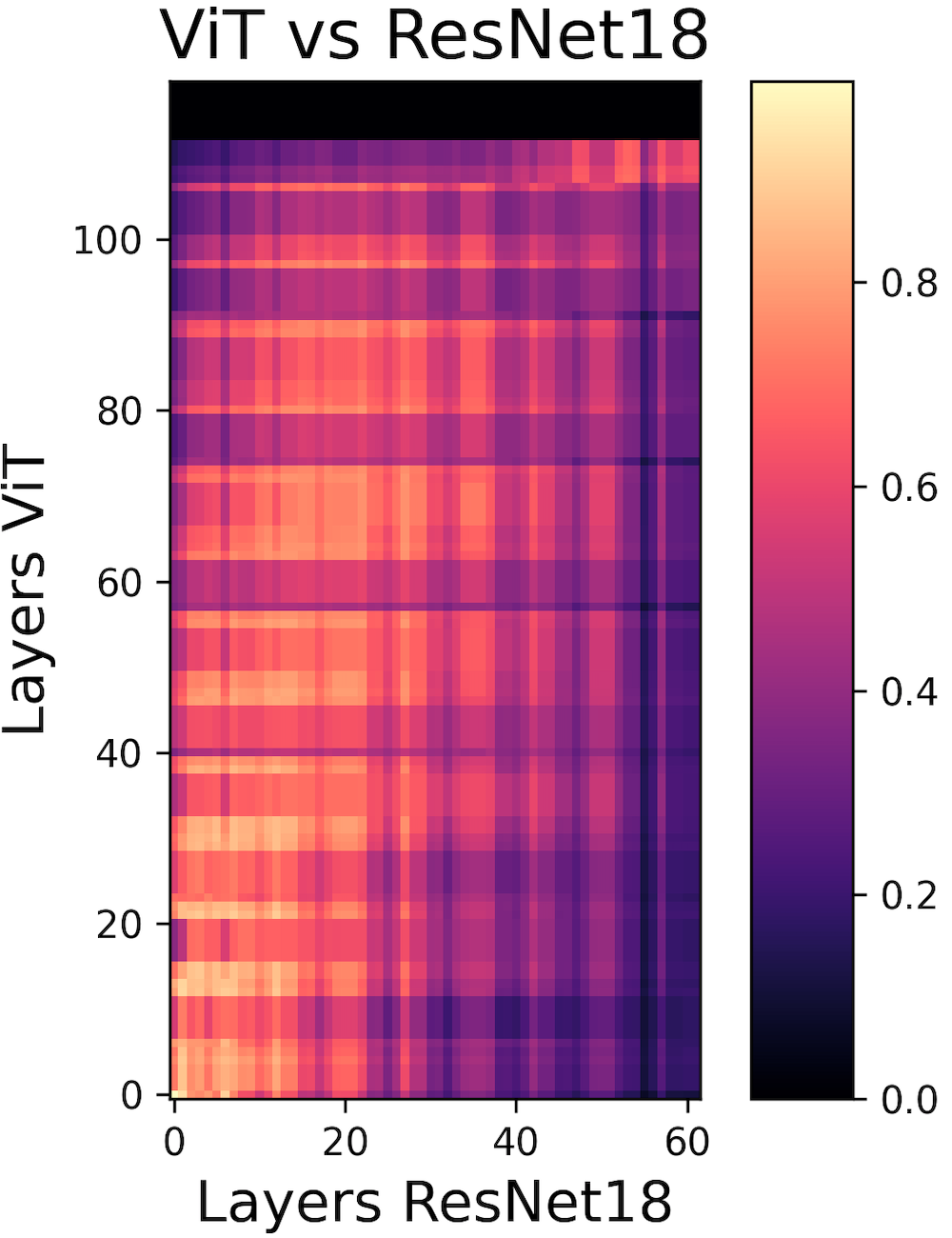}
     \caption{Representation similarity between ViT and ResNet on CIFAR-10}
     \label{fig:cka_cifar10}
 \end{subfigure}
 \hfill
 \begin{subfigure}[t]{0.3\textwidth}
     \centering
     \includegraphics[width=\textwidth]{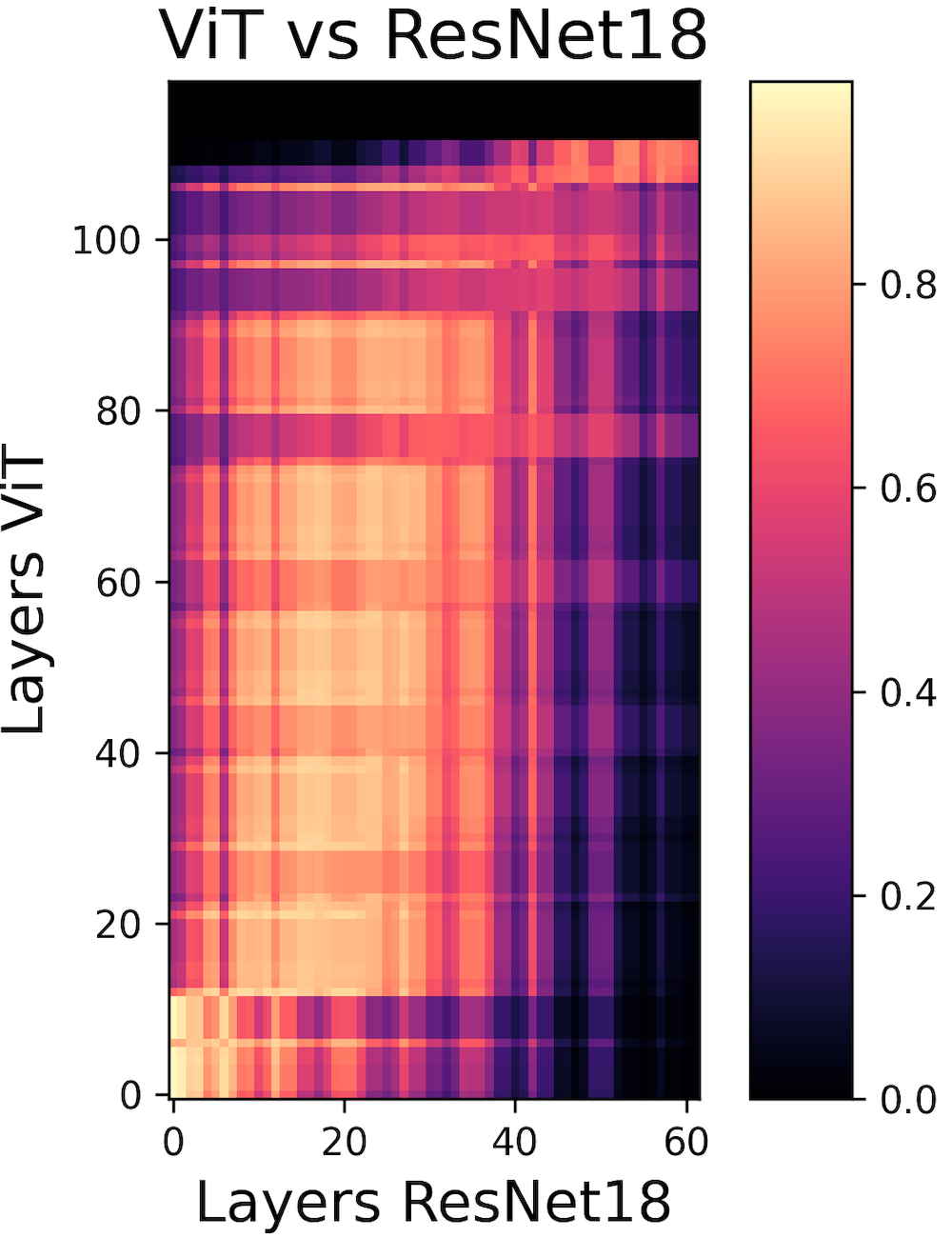}
     \caption{Representation similarity between ViT and ResNet on SVHN}
     \label{fig:cka_svhn}
 \end{subfigure}
    \caption{Representation similarity between ViT and ResNet on different datasets}
    \label{fig:cka_3model}
\end{figure}

After confirming our hypothesis that ViT performs less well compared to ResNet on small image datasets, we next try to provide an intuitive explanation on ViT's behavior when trained on a small dataset. We use \textbf{CKA (Centered Kernel Alignment)}\cite{raghu2021do} to analyze representation similarity between ViT and ResNet:
$$
\operatorname{CKA}(\boldsymbol{K}, \boldsymbol{L})=\frac{\operatorname{HSIC}(\boldsymbol{K}, \boldsymbol{L})}{\sqrt{\operatorname{HSIC}(\boldsymbol{K}, \boldsymbol{K}) \operatorname{HSIC}(\boldsymbol{L}, \boldsymbol{L})}}
$$
where $\mathbf{X} \in \mathbb{R}^{m \times p_{1}}$ and $\mathbf{Y} \in \mathbb{R}^{m \times p_{2}}$ are representations of two layers with $p_{1}$ and $p_{2}$, $\boldsymbol{K}=\boldsymbol{X} \boldsymbol{X}^{\top}$ and $\boldsymbol{L}=\boldsymbol{Y} \boldsymbol{Y}^{\mathrm{\top}}$ denote the Gram matrix for two layers, and HSIC is the Hilbert-Schmidt independence criterion \cite{Arthur2007}. In general, when the CKA value between two layers are high, the representations of these two layers are much similar. Using this metric, we can analyze the representation difference when ViT faces small datasets. Unlike \cite{raghu2021do} (See Figure \ref{fig:cka_jft_300m}), we focus on comparing the difference and giving interpretations on small datasets, which is novel. (See Figure \ref{fig:cka_cifar100}, Figure \ref{fig:cka_cifar10} and Figure \ref{fig:cka_svhn})

Note that when computing the representation similarity, we not only compare the convolution layers and attention layers, we also compare all the normalization and activation layers. By comparing Figure \ref{fig:cka_jft_300m} (from \cite{raghu2021do}, computed on large dataset) with Figure \ref{fig:cka_cifar100}, \ref{fig:cka_cifar10} and \ref{fig:cka_svhn} (computed on small datasets by us), we observe a huge change of representation power of ViT when trained from large datasets to huge datasets. From Figure \ref{fig:cka_jft_300m}, when trained with large datasets, almost the lower half of ResNet layers have a similar representation of the lowest quarter of layers in ViT; the latter half of ResNet layers have a similar representation of the next quarter of layers in ViT; the highest quarter of layers are dissimilar with all layers of ResNet. 

However, when trained on small datasets, the patterns of representation similarity change a lot. By comparing the representation similarity and observing the visualizations in the previous section, we observe: 

Figure \ref{fig:cka_cifar100} and Figure \ref{fig:cka_cifar10} change a lot, we completly lose the pattern on large datasets of Figure \ref{fig:cka_jft_300m}. Which means the representation on CIFAR-10 and CIFAR-100 is completely different with large datasets and causes the huge drop on final performance. 

Figure \ref{fig:cka_svhn} is most similar to Figure \ref{fig:cka_jft_300m}, we can still observe the lower layers of ResNet have similar representation of lower layers of ViT, higher layers of ResNet have similar representaion of middle-to-higher layers of ViT and highest layers of ViT are dissimilar with all layers of ResNet. However, in this case, ViT needs more layers to get the same representation of ResNet, compared with less layers before. 
From Figure \ref{fig:ViT_vis_plane}, lower layers of ViT is more focusing on local areas on SVHN dataset, which means that ViT can learn more locality on SVHN compared with CIFAR-10 and CIFAR-100. The reason might be that SVHN is a simpler dataset, thus ViT can catch the inductive bias of locality on this simple dataset and that can explain the reason why the performance of ViT is similar to ResNet on SVHN while it loses a lot on CIFAR-10 and CIFAR-100 in Table \ref{tab:accuracy_table}.

%% file: conclusion.tex
\section{Conclusion}

In this project, we would like to explore the reason why vision transformer does not perform well on small datasets. We firstly conduct extensive experiments to confirm the phenomenon of performance drop of ViT on small datasets. We later interpret the results by showing attention visualization and feature map visualization. Next, we conduct representation similarity analysis to further investigate the results. Finally, by comparing the difference between attention map visualization and representation similarity. We can speculate the reason for the performance drop of ViT on small datasets as follows:
\begin{itemize}
    \item When trained with small datasets, the representation of ViT is hugely different from ViT trained with large datasets and thus affects the performance a lot.
    \item The huge change of representation may be due to a lack of inductive bias of locality for ViT. Lower layers of ViT can not learn the local relations well with a small amount of data on complicated small datasets, e.g., CIFAR-10 and CIFAR-100. For simpler datasets, e.g., SVHN, ViT can learn locality relatively well, it's reflected on feature map visualization and might be the reason that ViT can achieve worse but similar performance on the SVHN dataset. 
\end{itemize}

%% file: acknowledgement.tex
\section{Acknowledgements}

Our code for visual transformer model construction and training is adapted from:
\begin{itemize}
    \item \url{https://github.com/kentaroy47/vision-transformers-cifar10}
\end{itemize}

CKA Similarity Analysis is adapted from: 

\begin{itemize}
    \item \url{https://github.com/AntixK/PyTorch-Model-Compare}
\end{itemize}

ViT attention weights extraction and visualization are adapted from:
\begin{itemize}
    \item \url{https://github.com/jeonsworld/ViT-pytorch/blob/main/visualize_attention_map.ipynb}
\end{itemize}

ResNet feature map visualization is adapted from:
\begin{itemize}
\item \url{https://ravivaishnav20.medium.com/visualizing-feature-maps-using-pytorch-12a48cd1e573}
\end{itemize}

%% file: appendix.tex
\section*{Appendix}

\subsection* {Additional Visualization Results}

\begin{figure}[h]
\centering
\includegraphics[width=15cm]{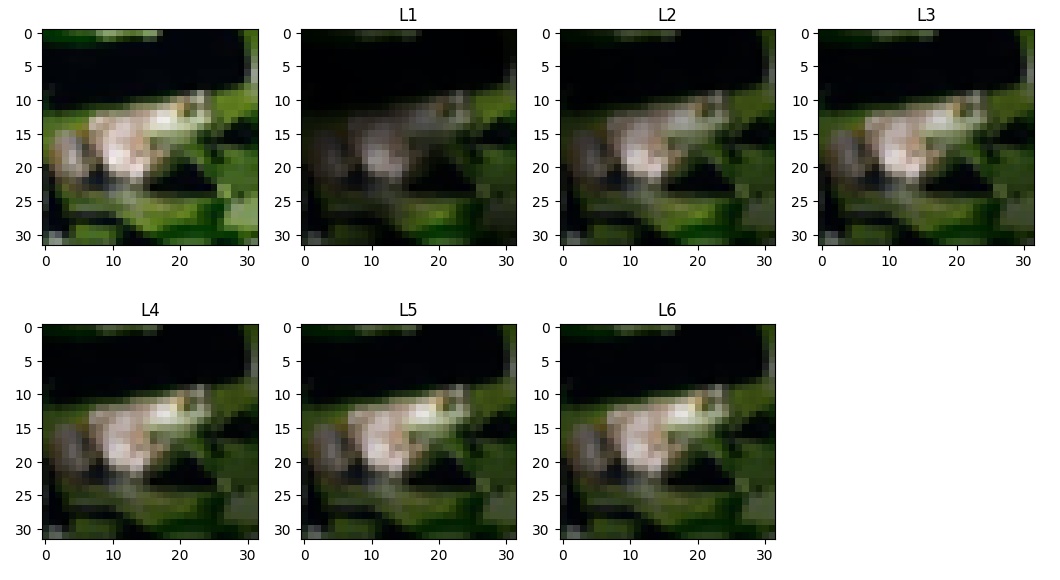} \\
\includegraphics[width=15cm]{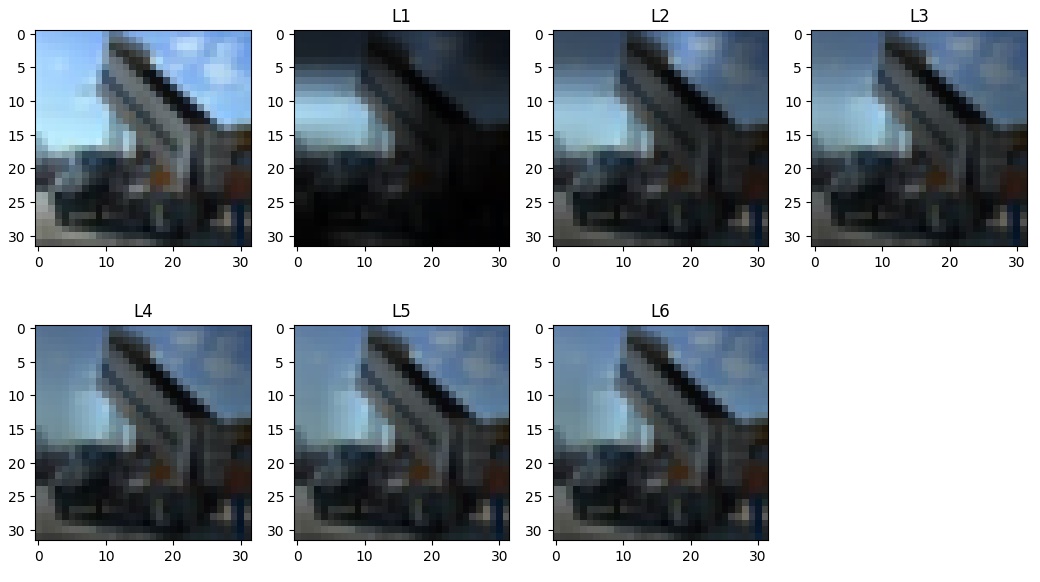} 
\caption{Visualization for ViT on CIFAR-10 dataset. }
\label{fig:ViT_vis_appendix}
\end{figure}

%% file: paper.bbl
\begin{thebibliography}{1}

\bibitem{ViT}
Alexey Dosovitskiy, Lucas Beyer, Alexander Kolesnikov, Dirk Weissenborn,
  Xiaohua Zhai, Thomas Unterthiner, Mostafa Dehghani, Matthias Minderer, Georg
  Heigold, Sylvain Gelly, Jakob Uszkoreit, and Neil Houlsby.
\newblock An image is worth 16x16 words: Transformers for image recognition at
  scale.
\newblock {\em CoRR}, abs/2010.11929, 2020.

\bibitem{naimi2021hybrid}
Safwen Naimi, Rien van Leeuwen, Wided Souidene, and Slim~Ben Saoud.
\newblock Hybrid byol-vit: Efficient approach to deal with small datasets.
\newblock {\em arXiv preprint arXiv:2111.04845}, 2021.

\bibitem{ViT_for_small}
Seung~Hoon Lee, Seunghyun Lee, and Byung~Cheol Song.
\newblock Vision transformer for small-size datasets.
\newblock {\em CoRR}, abs/2112.13492, 2021.

\bibitem{raghu2021do}
Maithra Raghu, Thomas Unterthiner, Simon Kornblith, Chiyuan Zhang, and Alexey
  Dosovitskiy.
\newblock Do vision transformers see like convolutional neural networks?
\newblock In A.~Beygelzimer, Y.~Dauphin, P.~Liang, and J.~Wortman Vaughan,
  editors, {\em Advances in Neural Information Processing Systems}, 2021.

\bibitem{netzer2011reading}
Yuval Netzer, Tao Wang, Adam Coates, Alessandro Bissacco, Bo~Wu, and Andrew~Y
  Ng.
\newblock Reading digits in natural images with unsupervised feature learning.
\newblock 2011.

\bibitem{liu2021efficient}
Yahui Liu, Enver Sangineto, Wei Bi, Nicu Sebe, Bruno Lepri, and Marco Nadai.
\newblock Efficient training of visual transformers with small datasets.
\newblock {\em Advances in Neural Information Processing Systems}, 34, 2021.

\bibitem{wandb}
Lukas Biewald.
\newblock Experiment tracking with weights and biases.
\newblock {\em Software available from wandb. com}, 2, 2020.

\bibitem{Arthur2007}
Arthur Gretton, Kenji Fukumizu, Choon Teo, Le~Song, Bernhard Sch\"{o}lkopf, and
  Alex Smola.
\newblock A kernel statistical test of independence.
\newblock In J.~Platt, D.~Koller, Y.~Singer, and S.~Roweis, editors, {\em
  Advances in Neural Information Processing Systems}, volume~20. Curran
  Associates, Inc., 2007.

\end{thebibliography}
